\documentclass{article}

\usepackage[final]{neurips_2024}

\usepackage[utf8]{inputenc} 
\usepackage[T1]{fontenc}    
\usepackage{hyperref}       
\usepackage{url}            
\usepackage{booktabs}       
\usepackage{amsfonts}       
\usepackage{nicefrac}       
\usepackage{microtype}      
\usepackage{xcolor}         
\usepackage{amsmath} 
\usepackage{graphicx}
\usepackage{subfig}
\usepackage{tipa}
\usepackage{xcolor}

\begin{document}

\title{Probing self-attention in self-supervised speech models for cross-linguistic differences}

\author{%
   Sai Gopinath \\
   Department of Computer Science\\
  University of Maryland, College Park\\
  \texttt{sgopinat@umd.edu} \\
    \And
  Joselyn Rodriguez \\
 Department of Linguistics \\
  University of Maryland, College Park \\
  \texttt{jrodri20@umd.edu} \\
}

\maketitle

\begin{abstract}
Speech models have gained traction thanks to increase in accuracy from novel transformer architectures (e.g., \cite{baevski_wav2vec_2020, mohamed_self-supervised_2022}.). While this impressive increase in performance across automatic speech recognition (ASR) benchmarks is noteworthy, there is still much that is unknown about the use of attention mechanisms for speech-related tasks. For example, while it is assumed that these models are learning language-independent (i.e., universal) speech representations, there has not yet been an in-depth exploration of what it would mean for the models to be language-independent. In the current paper, we explore this question within the realm of self-attention mechanisms of one small self-supervised speech transformer model (TERA). We find that even with a small model, the attention heads learned are diverse ranging from almost entirely diagonal to almost entirely global \textit{regardless} of the training language. We highlight some notable differences in attention patterns between Turkish and English and demonstrate that the models do learn important phonological information during pretraining. We also present a head ablation study which shows that models across languages primarily rely on diagonal heads to classify phonemes.
\end{abstract}

\section*{Introduction}
Most speech models utilized in automatic speech recognition (ASR) make use of unsupervised learning or self-supervised learning (SSL) algorithms as a lack of labeled training data makes supervised learning infeasible in most situations. Embeddings extracted from these pre-trained models can then be further fine-tuned on a variety of downstream tasks such as speech recognition, speaker identification, sentiment analysis, etc \cite{mohamed_self-supervised_2022}. In practice, these models have shown to be quite successful thanks to the models learning abstract representations that allow for application across other domains or languages (e.g., \cite{liu_tera_2021}). However, what information is actually contained within the learned representations of these SSL speech models is still under-explored. Previous findings so far have primarily shown that the embeddings extracted from SSL models display different acoustic or linguistic properties. For example, Pasad and colleagues probed embeddings at different layers of wav2vec 2.0 \cite{baevski_wav2vec_2020} finding that categorization accuracy differed across the layers of the network \cite{pasad_comparative_2022}. Additionally studies have explored whether these embeddings contain specific linguistic properties finding that speech embeddings contain both phonetic (low-level acoustic) information and phonemic (more abstract categorical) information \cite{martin_probing_2023}. However, this previous work has focused almost entirely on monolingual English-trained models. Whether there are cross-linguistic differences in these models, and what downstream consequences this might have (for example on multi-lingual fine-tuning) is under-explored. This is of importance in speech processing as not all languages make the same use of possible phonetic space (for example, the number of vowels in a language's phonology), or the kind of allowed phonotactics, or in the kind of phonological processes impacting the surface realization of the language. For example, Turkish contains a phonological process in which certain suffixes must agree within a word according to certain phonetic features (i.e., 'frontness' or 'roundess' of a vowel). This is shown in the example below:

\begin{table}
    \begin{center}
    \begin{tabular}{c|c|c}
        Nom Sg & Gen Sg & meaning \\
        \hline 
        pul & pul - un & 'stamp'\\
        el & el - in  & 'hand'
    \end{tabular}
        \caption{A simple example of vowel harmony in Turkish (from \cite{polgardi_vowel_1999}:}  
    \end{center}
\end{table}

In Turkish, /-in/ is the genetive suffix and shows two effects of "harmony" in which it must match in both roundness and frontness. Since [u] in "pul" is a back rounded vowel, the front high vowel [i] in /-in/ becomes [u] to match. Since /e/ is fronted and not rounded /-in/ surfaces as [-in]. Therefore, there is a strong relationship between phonological features of vowels in words in Turkish.

In this work, however, we show that regardless of cross-linguistic differences in a language's phonology, the training language does not necessarily impact the way in which SSL speech models encode information. Across three experiments, we find little differences between attention patterns learned for languages from two separate language families: Turkish, a a Turkic language which utilizes long-distance phonological patterns in the form of vowel harmony, and English, a Germanic language which does not.

\section*{Related Work}

Attention heads are key components that determine the significance of different parts of the input data by allocating varying degrees of attention or weight. This mechanism is central to the model's ability to capture dependencies and contextual relationships within the data, making it a critical area for analysis to understand model interpretability, optimize performance, and enhance transparency in decision-making processes.

One insight from attention head probing in natural language processing (NLP) is that self-attention mechanisms within text transformer models capture semantic content in addition to syntactic information, particularly in the context of sentiment analysis. Using a novel Layer-wise Attention Tracing method, it was shown that structured attention weights significantly contributed to identifying emotional semantics across different datasets \cite{wu_structured_2020}. This finding underscores the versatility and depth of information encoded in attention weights, suggesting that they can be pivotal in enhancing model interpretability and in drawing closer parallels with human cognitive processes in understanding language and emotions.

Of great relevance, another goal of this realm has been to determine which attention positions or heads are least useful so that they can be pruned to reduce computation without compromising output quality. For example, recent work has found that diagonal elements in the attention matrix are of less importance, suggesting their removal doesn't harm performance. This led to the development of SparseBERT, using a Differentiable Attention Mask (DAM) algorithm to optimize attention allocation \cite{shi_sparsebert_2021}.

While there has been a significant amount of research exploring the representations learned in text models, much less is known with respect to speech representations\cite{yang_understanding_2020}. This is surprising given that in many instances speech may contain long-distance phonological dependencies as discussed above, thus the efficacy of self-attention in specifically speech is unclear. This is crucial as not every language makes the same use of long-distance dependencies in phonological contexts. Understanding where or if cross-linguistic differences occur in the learned embeddings is important especially if the goal is learning language-agnostic representations while also optimizing model architecture and training. 

In one of the few works on attention mechanisms in speech transformers, Yang and colleagues ranked attention heads by type and pruned them to determine head importance, finding that for speech, global heads were \textbf{least} important, and -- surprisingly given results from research in NLP -- found that diagonal heads were most important. However, as stated above, different languages may place different importance on long distance relationships in the phonology, and thus while this work suggested that “global” heads are less important and can be safely pruned, this is not necessarily the case for languages like Turkish.

\section*{Model Architecture \& Training} 
The experiments were carried out using a medium-sized transformer model, TERA ('Transformer Encoder Representations from Alteration' an extension of Mockingjay; \cite{liu_tera_2021}) implemented through the sp3rl framework. TERA is a small self-supervised transformer model pre-trained using reconstruction loss where the model is tasked to minimize the reconstruction error of acoustic features when given an altered frame. There are three possible alterations, including frequency, magnitude, and time alterations. We use the base model (Layers=3, Parameters=21.3M). 

Two models were trained, one on English (tera-eng) and one one Turkish (tera-tur) speech corpora. The Turkish model was trained on approximately 100 hours of Turkish from the CommonVoice corpus \footnote{https://commonvoice.mozilla.org/en/datasets} for 200k steps with Adam optimizer while the English model is available pre-trained through the s3prl toolbox\footnote{https://github.com/s3prl/s3prl} on either 100 or 960 hour subsets of the librispeech\footnote{https://www.openslr.org/12} corpus of English. All models were pretrained using the spectogram reconstruction task, which involves recreating masked portions of the audio signal. Downstream training for each language was done on subsets of the same data used for pretraining the respective language. 

\section*{Experiment 1: Head classification and visualization of attention matrices}
\subsection*{Methods}
Generally, attention heads fall into three broad categories; global, vertical, and diagonal. Global attentions indicate that attention between different frames is random, vertical attentions indicate that specific frames are paid attention to across frames, and diagonal attentions indicate that frames mostly pay attention to themselves. 

While other attention head types have been explored in NLP (\cite{dark_bert}), these are the most comprehensive categories yet to be considered in speech transformers and are therefore also used in the current study for the sake of simplicity. When visualizing the heads, we did observe that most of them fall reasonably into these categories, however there were a few intermediate types (for example, block-diagonal, or vertical-diagonal). Future work will explore more fine-grained differences between head types.

Following \cite{yang_understanding_2020}, we use the following formulas to score the globalness, verticality, and diagonalness of all attention heads. In order to determine individual scores, we average across 10 random utterances for Turkish and English respectively.

In the formulas below (1-3), $\mathbb{H}$ refers to entropy and $u$ refers to each utterance in the total set of utterances $U$. $A^{h}$ refers to a single attention head; $A^{h}_{u}$ refers to the attention head for a given utterance, and $q$ and $k$ are the row and column indices of this attention head. $T$ is the number of frames for the given utterance, such that $q$ and $k$ are defined from 1 to $T$.

\begin{small}
\begin{align}
    G(A^{h}) = \mathbb{E}_{ u \sim U} \left[ \frac{1}{T} \sum^{T}_{q=1} \mathbb{H} (A^{h}_{u} [q]) \right]
\end{align}

\begin{align}
    D(A^{h}) = \mathbb{E}_{ u \sim U} \left[ -\frac{1}{T^{2}} \sum^{T}_{k=1}  | q - k | \cdot  A^{h}_{u} [q, k] \right]
\end{align}

\begin{align}
     V(A^{h}) = \mathbb{E}_{ u \sim U} \left[ -\mathbb{H}( \frac{1}{T} \sum^{T}_{q=1} A^{h}_{u} [q] )\right]
\end{align}
\end{small}

In all three formulas, the expectation of the value is taken over the distribution of utterances. The global formula computes the average entropy $H$ of each row $q$-indexed of the attention matrix $A^h_u$. Higher entropy indicates a more dispersed attention, implying less focus on specific areas. Intuitively, the diagonal formula can be taken as penalizing each value in the attention matrix the farther it is from the center diagonal line. In the vertical formula, we take the negative entropy of the average of each row in the attention matrix, penalizing high entropy as vertical heads should focus on specific individual targets across all positions. 

\subsection*{Results}
We categorize the attention heads for the Turkish pre-trained model and English pre-trained model separately. Examples of each head type for each language can be seen below in Figures 1 and 2. Note that the utterance lengths are longer for English than for Turkish.

\begin{figure}[!h]
\centering
    \subfloat[\centering Diagonal Head]{{\includegraphics[width=5cm]{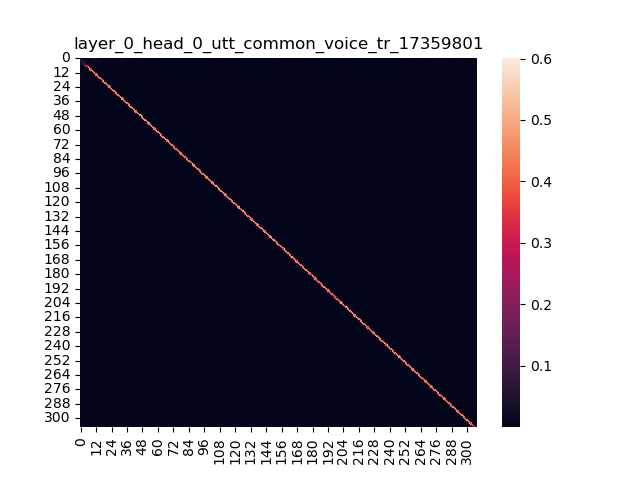} }}%
    \subfloat[\centering Vertical Head]{{\includegraphics[width=5cm]{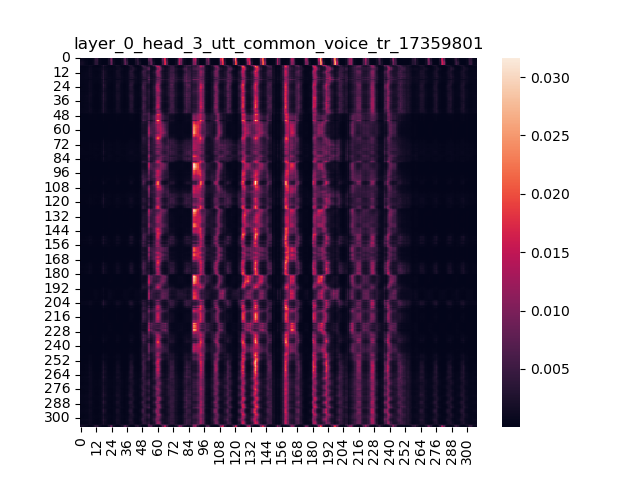} }}%
    \subfloat[\centering Global Head]{{\includegraphics[width=5cm]{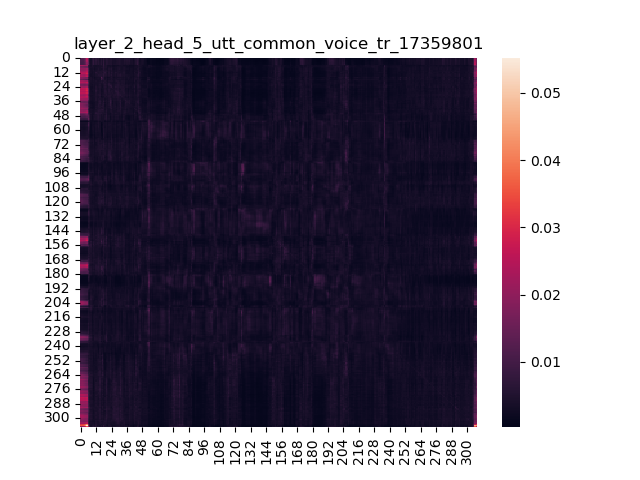} }}%
    \caption{Examples of single Turkish utterance heads from each category for Turkish pre-trained TERA. Both axis correspond to frames.}%
    \label{fig:example}%
\end{figure}
\begin{figure}[!h]
\centering
    \subfloat[\centering Diagonal Head]{{\includegraphics[width=4.5cm]{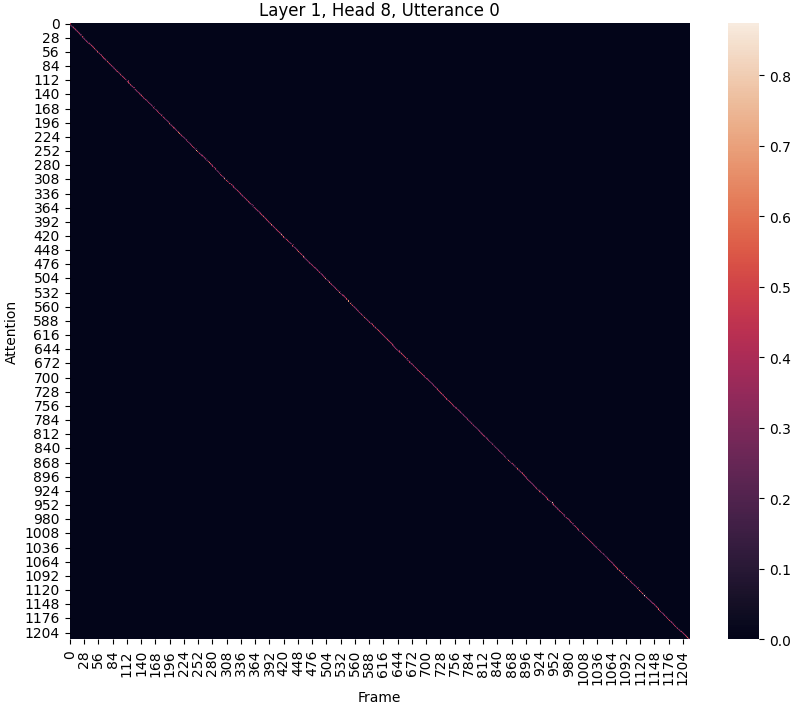} }}%
    \subfloat[\centering Vertical Head]{{\includegraphics[width=4.5cm]{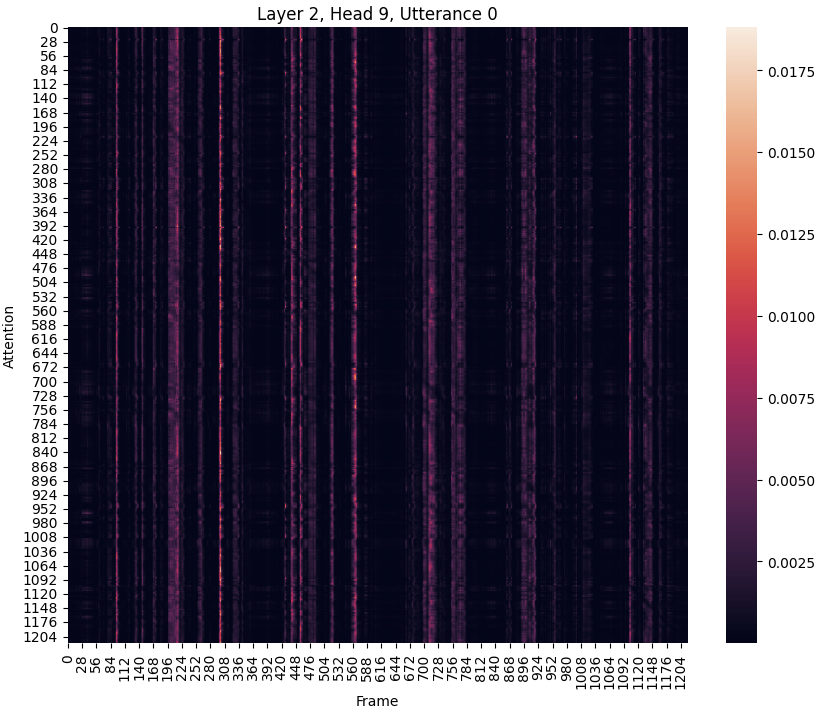} }}%
    \subfloat[\centering Global Head]{{\includegraphics[width=4.5cm]{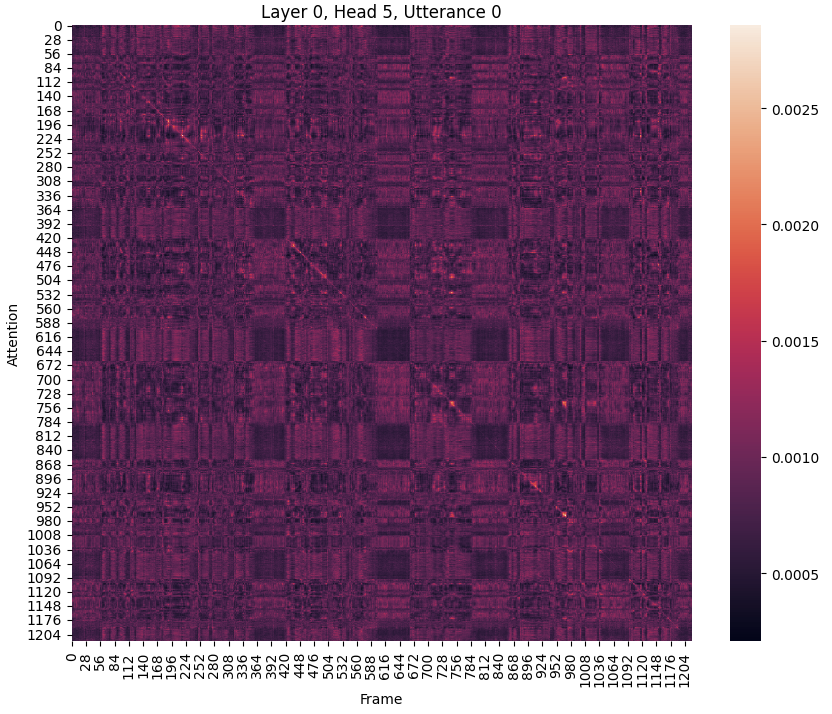} }}%
    \caption{Examples of single English utterance heads from each category for English pre-trained TERA. Both axis correspond to frames.}%
    \label{fig:example}%
\end{figure}

The difference in scores for each formula for all heads can be seen in Table 2. The difference in number of heads in each category for Turkish and English can be seen in Figure 3.

\begin{table}[ht]
\centering
\begin{tabular}{|l|c|}
\hline
\textbf{Metric} & \textbf{Value} \\ \hline
Global English & 5.555434260059006 \\ \hline
Global Turkish & 4.488142476890746 \\ \hline
Vertical English & -6.919290277693065 \\ \hline
Vertical Turkish & -5.853183719846937 \\ \hline
Diagonal English & -0.11835675427250357 \\ \hline
Diagonal Turkish & -0.2504278951056385 \\ \hline
\end{tabular}
\caption{Average category scores for English and Turkish}
\label{tab:avg_scores}
\end{table}

We can see that the Turkish heads are on average less global, more vertical, and less diagonal.The largest difference in number of heads can be seen in the vertical category, where the Turkish model learns significantly more vertical heads than for English. 

\begin{figure}[!ht]
    \centering
    \includegraphics[width=.75\linewidth]{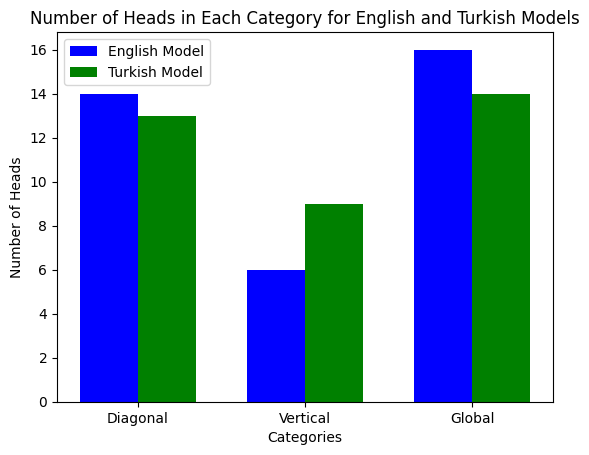}
    \caption{Number of Heads in Each Category for Each Language}
    \label{fig:barchart}
\end{figure}

\begin{figure}[!h]
\centering
    \subfloat[\centering Global Heads and Values]{{\includegraphics[width=5cm]{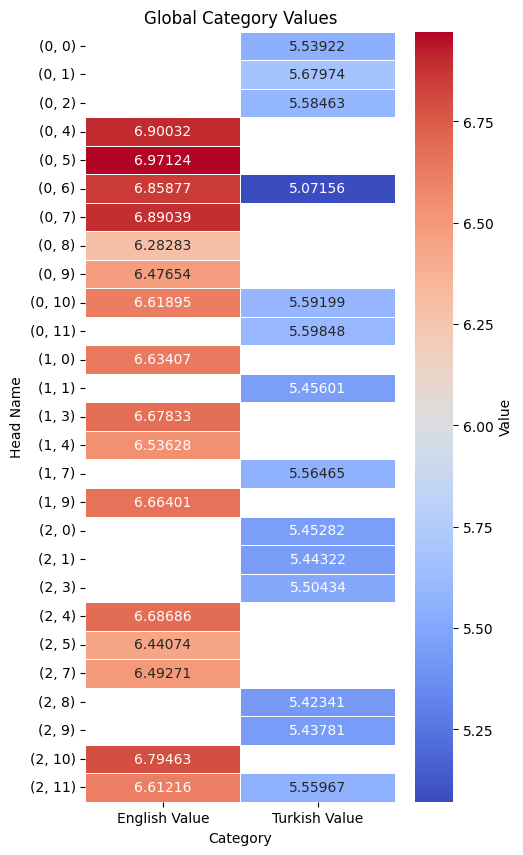} }}%
    \subfloat[\centering Vertical Heads and Values]{{\includegraphics[width=5cm]{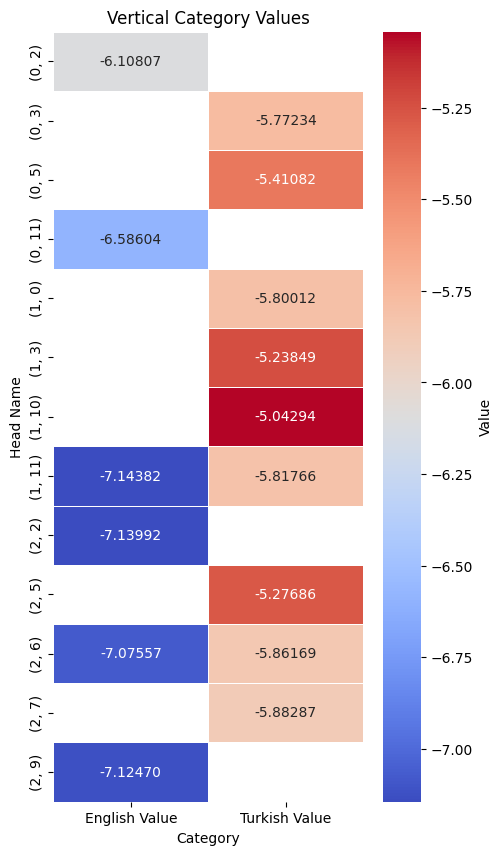} }}%
    \subfloat[\centering Diagonal Heads and Values]{{\includegraphics[width=5cm]{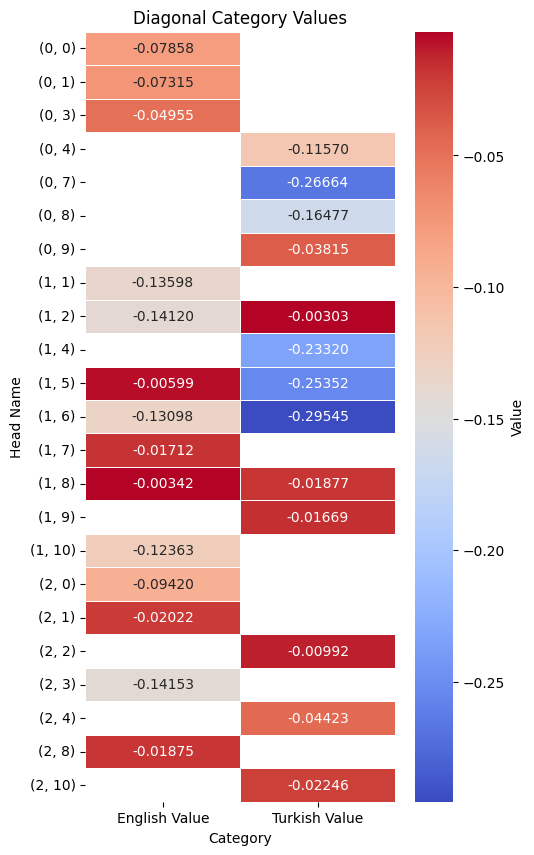} }}%
    \caption{English vs Turkish Heads and score values in each category. The y axis corresponds to the head name (layer number, head number), the columns are the languages, and the values are the scores for each head according to the category formulas. A blank space means that that head fell into a given category (diagonal, vertical, global) for one language but not the other}%
    \label{fig:heatmaps}%
\end{figure}

Differences in metric values between languages and head types across layers are shown below in Figures 4 where it can be seen that there are more diagonal heads concentrated with higher values in layers 1 and 2 than layer 0 for English. This pattern is not as discernible for Turkish. For global English heads, the strongest heads are concentrated in layer 0, but this is not the case for Turkish. This does not support the finding in \cite{shim_understanding_2022}, where they found that the earlier layers contained more diagonal heads. However, the model used in current work is significantly smaller, and therefore may not have as much difference across layers of the network. This is important to note that because of this, the results of the current work may be constrained by the size of the network. Given that diagonal heads are shown to be of greatest importance in phoneme classification in Experiment 3, it's possible that the fewer global heads for Turkish is partially responsible for the lower overall accuracy.

\section*{Experiment 2: Phoneme Relation Maps}

\subsection*{Methods}
In order to determine the relation between phonemes in a language, we constructed phoneme relation maps (PRM) relating the amount of attention a given phoneme attends to another phoneme, quantifying this difference as the average over all attention relations. Specifically, we average the attention  $A_{u}^{h}[k, q]$ between all phone pairs $Y_m$, $Y_n$ in $Y$, the set of all possible phones in the language for a given set of labels, $y$, for each utterance where $y_m = Y_m$ and $y_n = Y_n$ (see \cite{yang_understanding_2020} for more detail). Intuitively, this gives a measure of how often on average a given phoneme attends to another phoneme. Attention matrices for each utterance were extracted for both English and Turkish for all 3 layers of the model, which each contained 12 attention heads for a total of 36 attention matrices. Given the large number of heads, we report the average over all heads (n=12) in the last layer of the network (layer=3) for 200 utterances from English and Turkish respectively. 

\subsection*{Results}

The phoneme relation maps averaged over the final layer for Turkish and English are displayed below. As can be seen from both maps, there is not a clear distinction between the learned relations for Turkish and English. While it was hypothesized that Turkish may have higher relations between vowels due to the importance of vowel features in the language, this is not evident from the data. Rather, for both English and Turkish, phonemes tend to relate primarily to themselves. Additionally, it doesn't seem to be the case that vowel on general attend to to vowels or consonants to consonants. As can be seen from the vowels in the blue box, there is not a clear difference in the amount of attention between specific vowel phonemes as compared to consonant phonemes. 

\begin{figure}[!h]
\centering
    \includegraphics[width=.75\linewidth]{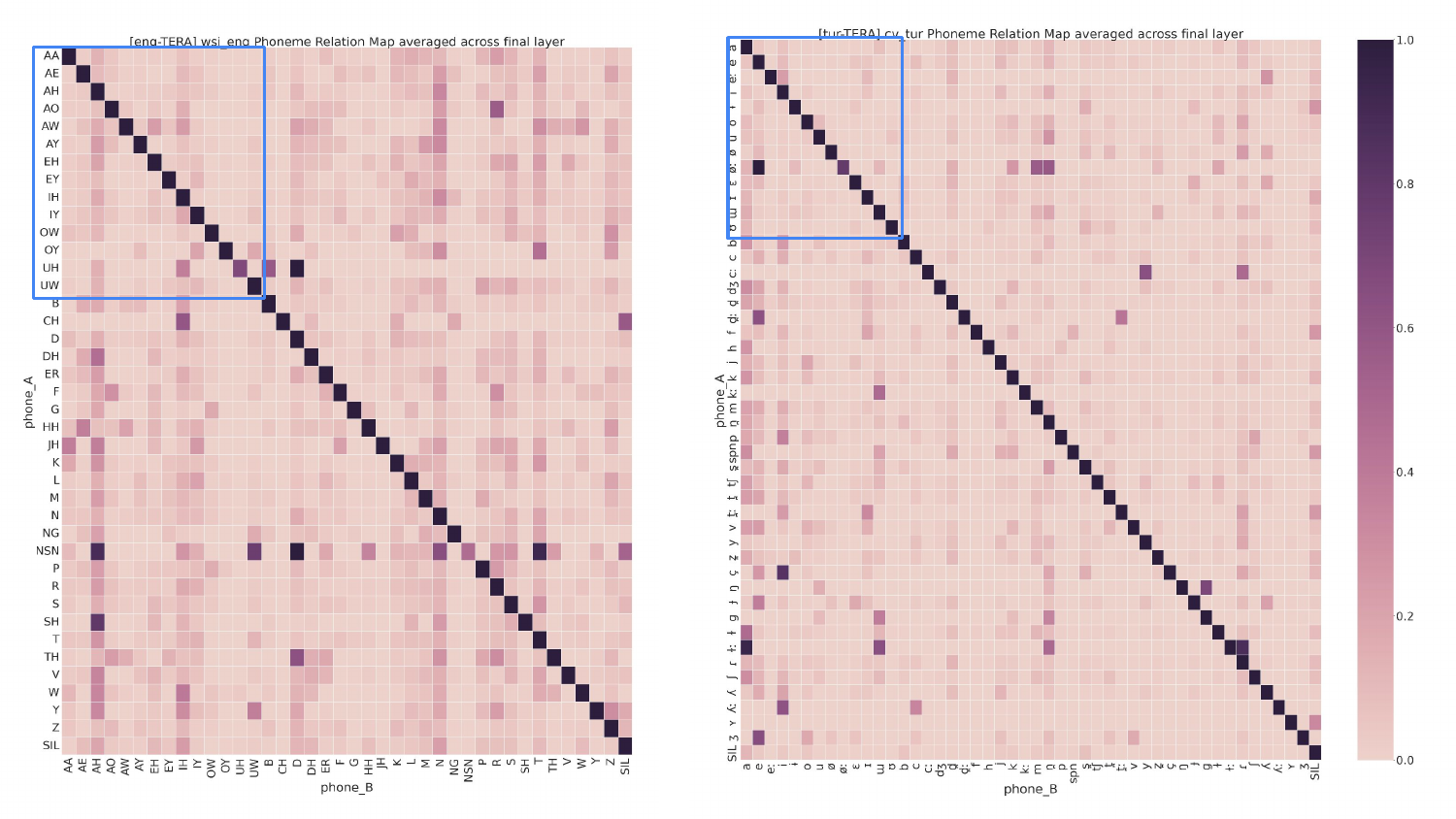}
    \caption{Phoneme relation maps for English (left) and Turkish (right) with vowels identified in blue box}%
    \label{fig:prm}%
\end{figure}

Rather, the relation is much more varied, possibly suggesting more fine-grained phonotactic influence (e.g., a specific vowel occur in specific phonetic contexts within the language). We leave exploring this avenue to future work. 

\section*{Experiment 3: Phoneme Classification Accuracy Between Languages and Categorically Masked Heads}
\subsection*{Finetuning Accuracy Between and Across Languages}
Phoneme classification is the task of predicting which phoneme, or a unit of sound, is being spoken at each given frame in the input. While there are different ways to break down sounds into phonemes, we rely on the pronunciation dictionary we use for English provided by s3prl that consists of 41 phonemes. We used a Turkish dictionary provided through the montreal forced aligner with a 48 phoneme classes (\url{https://mfa-models.readthedocs.io/en/latest/dictionary/Turkish/Turkish%20CV%20dictionary%20v2_0_0.html#Turkish%20CV%20dictionary%20v2_0_0}). 

To train a model on phoneme classification, we first created phoneme frame alignments. First, we obtain time-level alignments, where the phoneme start and end time is labeled in seconds. For Librispeech train-clean-100, we had access to alignments through s3prl, which contains high-quality hand-aligned phoneme aligments. For the Turkish data, we used the Montreal Forced Aligner, a toolkit for forced alignment providing pre-trained acoustic models, which was further adapted on the Turkish common voice data. During data preprocessing, we convert the phoneme start and end times into frame level alignments such that each individual frame is mapped to a phoneme. In TERA, a frame starts every 10 milliseconds with a 25 millisecond window. Generally phonemes will be spoken over the course of multiple frames. This becomes a linear classification task, with 41 classes for English and 48 classes for Turkish (including silence and unidentified), where a class is predicted for each frame. 

We trained a frame-level linear prediction model on top of our pretrained TERA model to predict phonemes. We used an 80/20 split between training and test data; the total was $\sim$20 hours of labeled data. For all experiments, we trained for 50k steps. Due to available data constraints, the training was done on data that was seen during pretraining, so the accuracy metric may not be a good measurement of real-world phoneme classification accuracy. 

We had three base models initially; an English model pretrained on 100 hours, another pretrained on 960 hours, and a Turkish model pretrained on 108 hours. The English and Turkish model which were trained on $\sim$100 hours were each trained for 200k steps, whereas the 960 hour pretrained English model was trained for 500k steps. Each model was finetuned on the English and Turkish phoneme classification task separately, and accuracy was evaluated on the held out test set. The results can be seen in Table 3. 

From the table, we can see that none of the models were able to break 50\% on the Turkish test data. This is likely due to the quality of alignments; there was much less data to train the Turkish force aligner than was used for the English alignments. Through visual inspection of the alignment in comparison to the Turkish force alignments, we were able to see that they were generally aligned well, but small errors in marking the exact beginning and ends of each phoneme could have a large effect when we look at the frame-level alignments, which start every 10 milliseconds. We can see that the Turkish model performs slightly better on Turkish than the 100 hour English model, which performs slightly better than the 960 hour model, indicating that the model did learn some useful phoneme information during Turkish pretraining, and learning more English specific patterns does hinder performance on Turkish. However, the effect size here is  small. 

On the other hand, the difference in performance between the English and Turkish models on English phonemes is vast. The difference on accuracy for the Turkish model on the Turkish phonemes vs English phonemes is also notable. This effectively illustrates the difference that the pretraining language makes, showing that the models indeed acquire language-specific information.  

\begin{table}[!ht]
\centering
\begin{tabular}{|c|c|c|c|c|}
\hline
\textbf{Pretrained Language} & \textbf{Pretrained hours} & \textbf{Finetuned Language} & \textbf{Finetuned hours} & \textbf{Test Accuracy} \\ \hline
English & 960 & Turkish & 20.3 & 0.4861824811 \\ \hline
English & 100 & Turkish & 20.3 & 0.4928628504 \\ \hline
English & 960 & English & 20.3 & 0.7289866209 \\ \hline
English & 100 & English & 20.3 & 0.7145705819 \\ \hline
Turkish & 108 & Turkish & 20.3 & 0.4990699291 \\ \hline
Turkish & 108 & English & 20.3 & 0.4752979279 \\ \hline
\end{tabular}
\caption{English and Turkish Language Model Accuracies on English and Turkish Phoneme Classification}
\label{tab:language_model_accuracies}
\end{table}

\subsection*{Head Ablation Study}
Next, we investigate the effects of different types of attention heads on Turkish and English performance. We use only the Turkish pretrained model for the Turkish task, and vice versa for English. The English model used was the one pretrained for 100 hours to be more comparable. We cumulatively mask heads in each category (global, vertical, and diagonal), starting with the highest ranked heads (the highest value from the formula corresponding to their categorization). The results can be seen in Figures 6 and 7.

\begin{figure}[!ht]
    \centering
    \includegraphics[width=.75\linewidth]{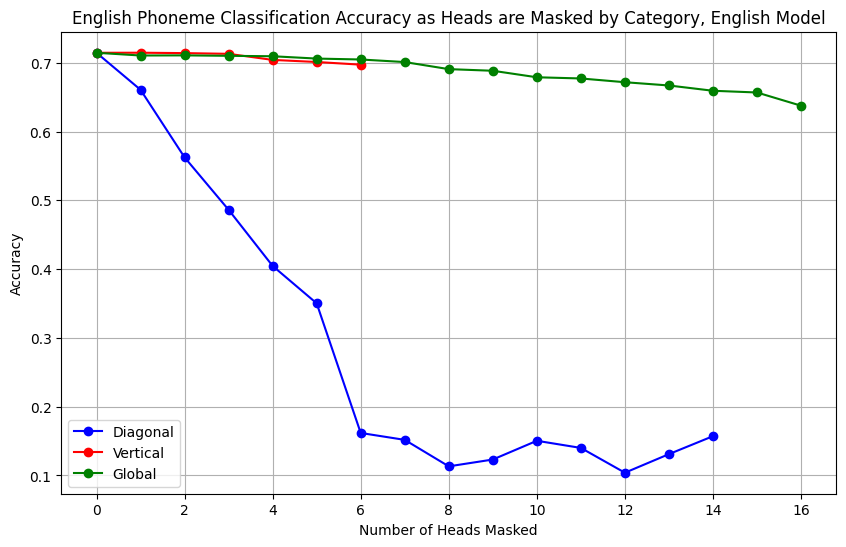}
    \caption{Head Ablation Study - English Model, English Task}
    \label{fig:engablation}
\end{figure}
\begin{figure}[!ht]
    \centering
    \includegraphics[width=.75\linewidth]{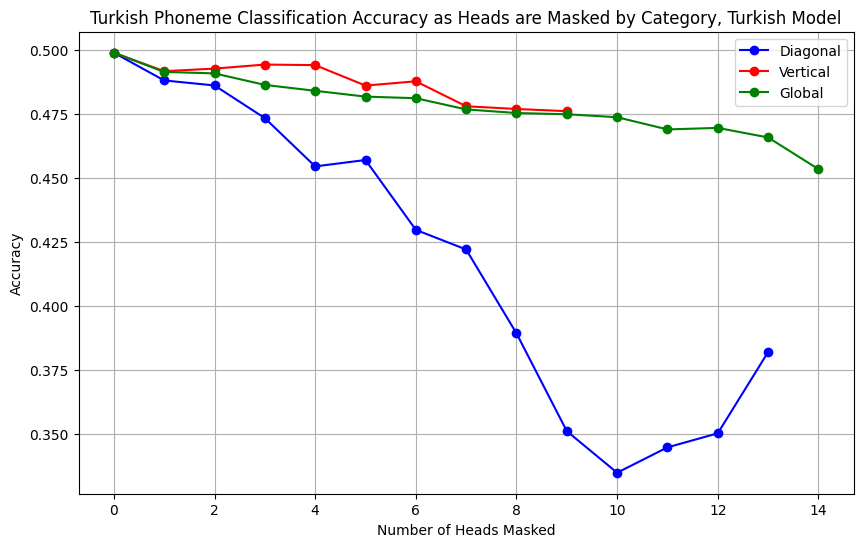}
    \caption{Head Ablation Study - Turkish Model, Turkish Task}
    \label{fig:turablation}
\end{figure}

To contextualize the results, the accuracy on Turkish and English with no attention heads was $\sim$44.6\% and $\sim$39\% respectively. In line with \cite{yang_understanding_2020}, there is a slight downward trend when removing vertical and global attention heads, but occasionally the accuracy slightly rises. Interestingly, the Turkish model learned more vertical attention heads during pretraining, but some of them might even be harmful for the task of phoneme classification. 

Also interestingly, when masking diagonal heads, the accuracy drops far below that of the model when all heads are masked for both languages. Though vertical and global heads as a whole seem to be slightly helpful for classifying phonemes, forcing the model to only rely on them drops the accuracy far below than if the model was not using attention at all. The slight increase in accuracy towards the end can likely be explained by the fact that the lower-ranking diagonal heads contained enough vertical and global attention to be harmful for classification without the higher-ranking diagonal heads. 

These results support the general assumption that models rely on diagonal heads to classify phonemes across languages. Some vertical and global heads can be pruned to increase efficiency, but this should be done with caution and with regards to specific downstream tasks. There do appear to be some differences in how different types of attention may be used between English and Turkish, but without comparable data quality it is difficult to make any strong conclusions.

\section*{Conclusion}
From fine-tuning on English and Turkish phoneme classification, we did gather empirical evidence that important phonological information is encoded by speech models. Most of the results were limited by noisy alignments on the Turkish data, however, the different in performance for the Turkish pretrained model and the English pretrained model on the English phoneme classification task is quite vast ($\sim$48\% and $\sim$71\% respectively). There was also a slight difference in accuracies between the Turkish model and English model on the Turkish task ($\sim$49.99\% versus $\sim$49.2\%). We speculate the difference was low because of the noisy test data for Turkish; it is possible that the maximum reasonable accuracy that can be achieved on the noisy data without overfitting is closed to 50\%. Given that the accuracies with all attention heads masked was quite low, it does seem like some language specific differences are encoded within the attention mechanism.

Despite this finding of language-specificity in phoneme categorization, we were unable to find satisfactory explanations for this performance gap within the attention heads. Given the lack of relation between phoneme attention relations between languages, our findings did not support the hypothesis that a Turkish model would learn more long-distance phonological dependencies. However, we note some interesting patterns between languages that were not fully explainable. Thus, there is a lot of room for future research in this area. Perhaps different methods for analyzing attention, for example more fine-grained categories, are needed. 

The results of our head ablation study on the English and Turkish models support the general finding that speech models rely on diagonal heads to segment and classify phonemes. Diagonal heads in both English-trained and Turkish-trained transformer models are of greatest importance in a phoneme classification task and both global and vertical heads can be safely pruned with significantly less consequence. This finding is significant, since it suggests that pruning strategies to reduce model overhead may actually be applicable cross-linguistically. In addition, we revealed that attention heads in both languages rely on each other, as masking only diagonal heads resulted in performance dropping below the fully masked baseline. 

The greatest limitations of our study and areas for future work to improve are the noisy Turkish data and limited availability in general of Turkish speech data. Future work may also expand the current attention analysis methods to more fine-grained categories. Further studies should take into consideration other methods, including using more categories to label attention heads. Additionally, the methodology of the paper should ideally be extended to other self-supervised transformer models trained using different tasks and languages to provide broader results.


\appendix

\section{Appendix}
\subsection{Supplemental material}
We have made our work available \href{https://github.com/joselyn-rodriguez/harmony}{here}, which contains code to extract, display, and categorize attention heads.

To conduct our experiments, it was necessary to modify the s3prl library to output attention heads, keys, and queries. We have made our modifications available \href{https://github.com/joselyn-rodriguez/s3prl}{here}.

\subsection{License Declarations}
Most of S3PRL is licensed under Apache License version 2.0, and a few files are licensed under CC-BY-NC.
Librispeech is licensed under CC BY 4.0.
The TERA model is licensed under Apache License version 2.0. 

\end{document}